\pdfoutput=1

\documentclass[10pt,conference,a4paper]{IEEEtran}

\ifCLASSINFOpdf
\else
\fi

\usepackage{graphicx}
\usepackage{amsmath,amssymb} 

\usepackage[american]{babel}
\usepackage{microtype}

\usepackage{bm}
\usepackage{graphicx}
\usepackage{subfigure}

\usepackage{makecell}
\usepackage{multirow}
\usepackage{url}

\DeclareMathOperator*{\argmax}{\arg\!\max}
\DeclareMathOperator{\logadd}{logadd}

\begin{document}

\title{PICK: Processing Key Information Extraction from Documents using Improved Graph Learning-Convolutional Networks}


\author{

\IEEEauthorblockN{\IEEEauthorrefmark{1}Wenwen Yu\IEEEauthorrefmark{2},
\IEEEauthorrefmark{1}Ning Lu\IEEEauthorrefmark{3},
Xianbiao Qi\IEEEauthorrefmark{3},
Ping Gong\IEEEauthorrefmark{2} and
Rong Xiao\IEEEauthorrefmark{3}}

\IEEEauthorblockA{\IEEEauthorrefmark{2}School of Medical Imaging, Xuzhou Medical University, Xuzhou, China
}

\IEEEauthorblockA{\IEEEauthorrefmark{3}Visual Computing Group, Ping An Property \& Casualty Insurance Company, Shenzhen, China\\
}

\IEEEauthorblockA{Email: yuwenwen62@gmail.com, gongping@xzhmu.edu.cn\\
\{jiangxiluning, qixianbiao, rongxiao\}@gmail.com\\
\IEEEauthorrefmark{1} \emph{denotes equal contribution}
}

}

\maketitle

\begin{abstract}
Computer vision with state-of-the-art deep learning models has achieved huge success in the field of Optical Character Recognition~(OCR) including text detection and recognition tasks recently. However, Key Information Extraction~(KIE) from documents as the downstream task of OCR, having a large number of use scenarios in real-world, remains a challenge because documents not only have textual features extracting from OCR systems but also have semantic visual features that are not fully exploited and play a critical role in KIE. Too little work has been devoted to efficiently make full use of both textual and visual features of the documents. In this paper, we introduce PICK, a framework that is effective and robust in handling complex documents layout for KIE by combining graph learning with graph convolution operation, yielding a richer semantic representation containing the textual and visual features and global layout without ambiguity. Extensive experiments on real-world datasets have been conducted to show that our method outperforms baselines methods by significant margins. Our code is available at https://github.com/wenwenyu/PICK-pytorch.
\end{abstract}

\IEEEpeerreviewmaketitle

\section{Introduction}
Computer vision technologies with state-of-the-art deep learning models have achieved huge success in the field of OCR including text detection and text recognition tasks recently. Nevertheless, KIE from documents as the downstream task of OCR, compared to typical OCR tasks, had been a largely under explored domain and is also a challenging task~\cite{huang2019icdar2019}. The aim of KIE is to extract texts of a number of key fields from given documents, and save the texts to structured documents. KIE is essential for a wide range of technologies such as efficient archiving, fast indexing, document analytics and so on, which has a pivotal role in many services and applications.

\begin{figure*}[htbp]
\centering
\subfigure[Medical invoice]{
\begin{minipage}[b]{0.3\textwidth}
\includegraphics[height=.6\textwidth]{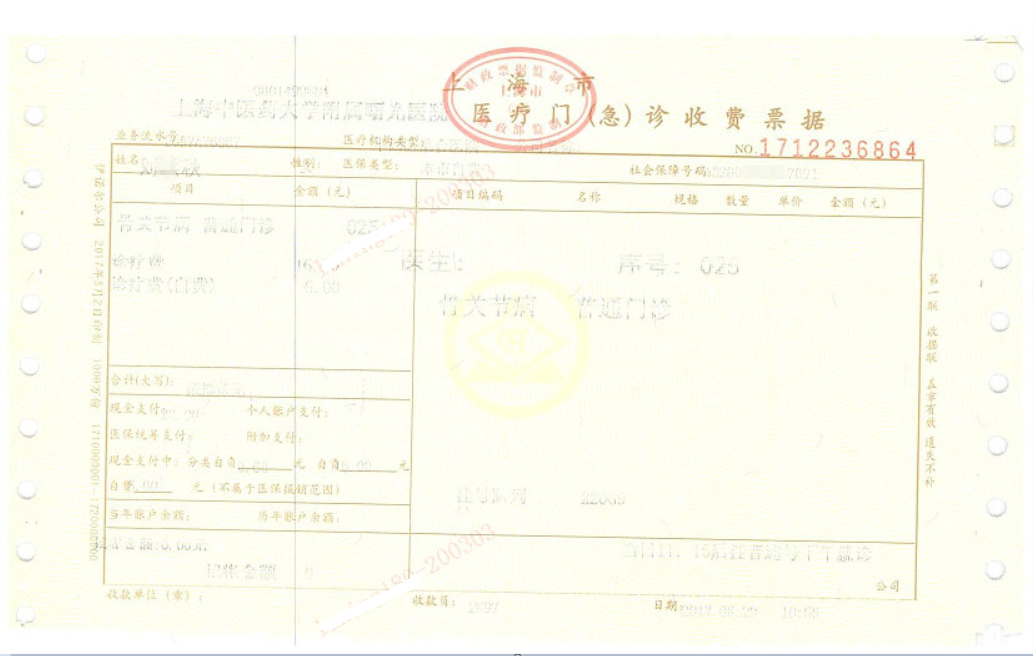}
\label{fig:medical-invoice}
\end{minipage}} 
\subfigure[Train ticket]{
\begin{minipage}[b]{0.3\textwidth}
\includegraphics[height=.6\textwidth]{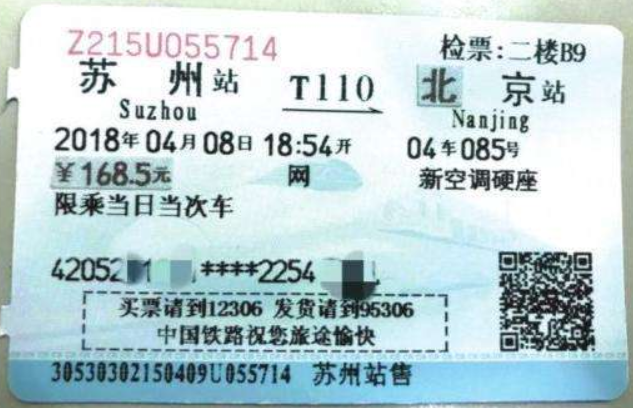}
\label{fig:eaten-trainticket}
\end{minipage}}
\subfigure[Tax receipt]{
\begin{minipage}[b]{0.3\textwidth}
\includegraphics[height=.6\textwidth]{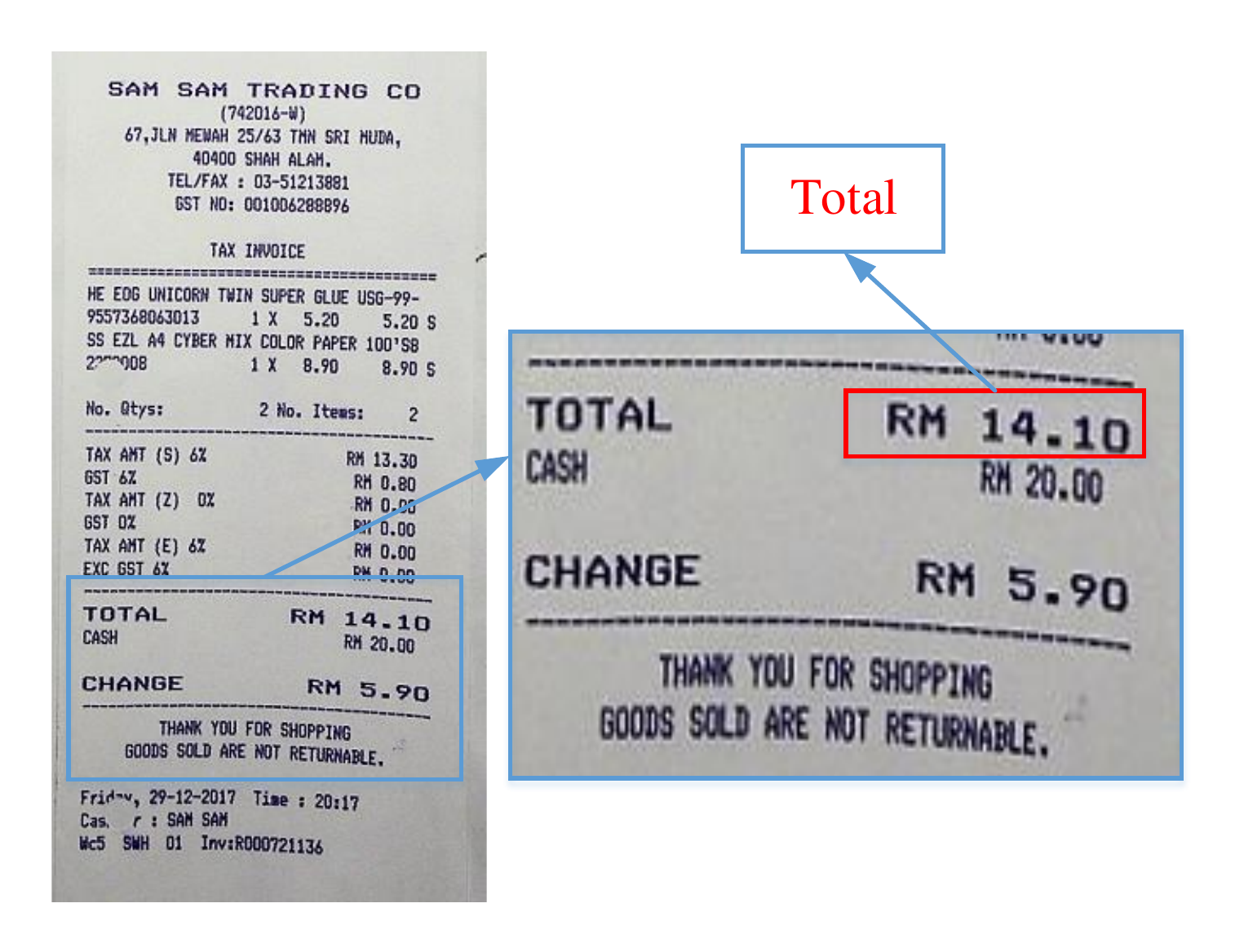}
\label{fig:sroie-invoice}
\end{minipage}}
\label{fig:examples}
\caption{Examples of documents with different layouts and types.}
\end{figure*}

Most KIE systems simply regard extraction tasks as a sequence tagging problems and implemented by Named Entity Recognition (NER)~\cite{lample2016neural} framework, processing the plain text as a linear sequence result in ignoring most of valuable visual and non-sequential information (e.g., text, position, layout, and image) of documents for KIE. The main challenge faced by many researchers is how to fully and efficiently exploit both textual and visual features of documents to get a richer semantic representation that is crucial for extracting key information without ambiguity in many cases and the expansibility of the method~\cite{aumann2006visual}. See for example Figure~\ref{fig:sroie-invoice}, in which the layout and visual features are crucial to discriminate the entity type of TOTAL. Figure~\ref{fig:examples} shows different layouts and types of documents.

\begin{figure}[htb]
\centering
\includegraphics[height=.4\textwidth]{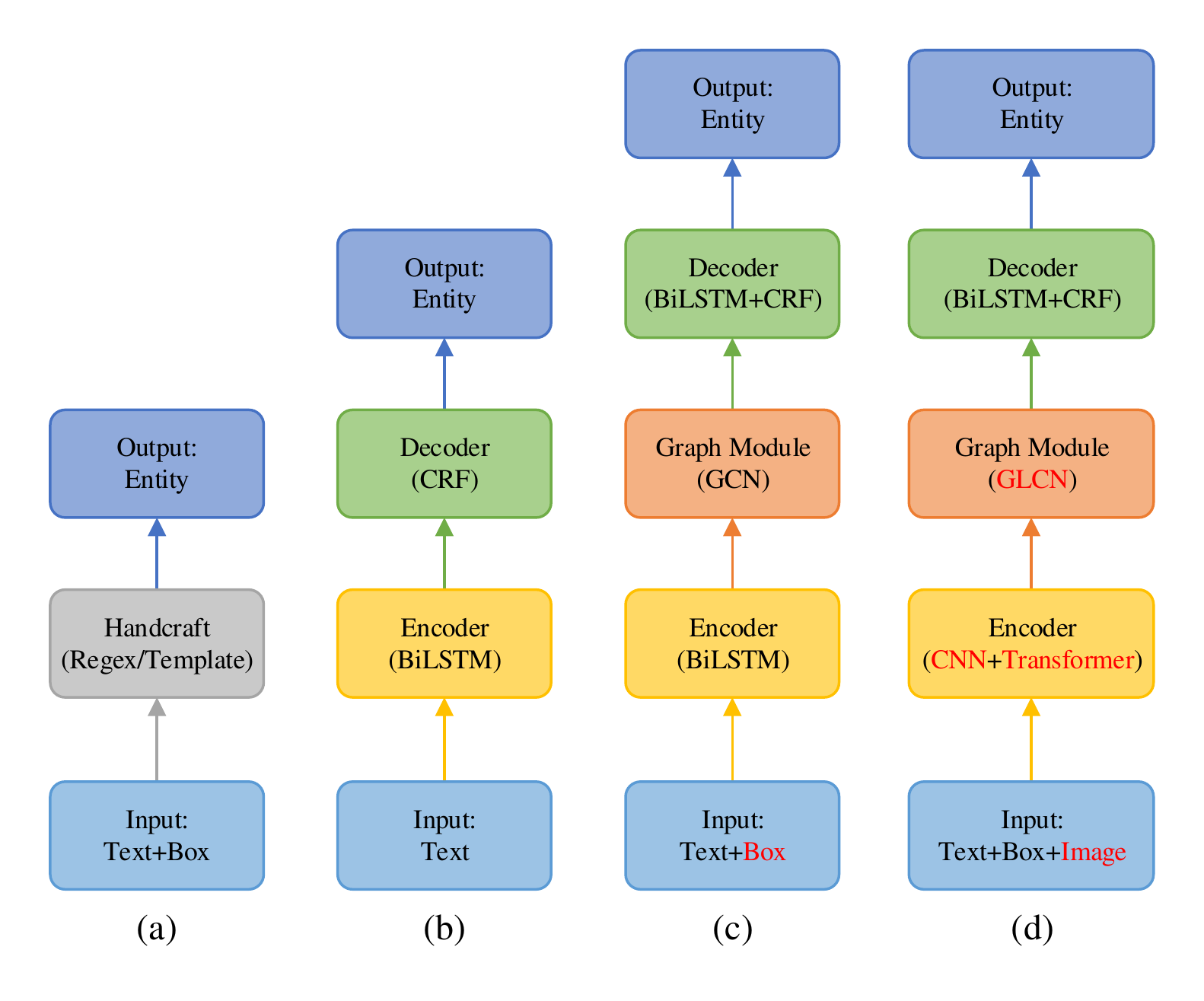}
\vspace{-1em}
\caption{Typical architectures and our method for key information extraction. (a)~hand-craft features based method.  (b)~automatic extraction features based method. (c)~using more richer features based method. (d) our proposed models.}
\label{fig:compare}
\end{figure}

The traditional approaches use hand-craft features (e.g., regex and template matching) to extract key information as shown in Figure~\ref{fig:compare}(a). However, this solution~\cite{schuster2013intellix,simon1997fast} only uses text and position information to extract entity and need a large amount of task-specific knowledge and human-designed rules, which does not extend to other types of documents. Most modern methods considered KIE as a sequence taggers problem and solved by NER as shown in Figure~\ref{fig:compare}(b). In comparison to the typical NER task, it is much more challenging to distinguish entity without ambiguity from complicated documents for a machine. One of the main reasons is that such a framework only operates on plain texts and not corporates visual information and global layout of documents to get a richer representation. Recently, a few studies in the task of KIE have attempted to make full use of untapped features in complex documents. \cite{xu2019layoutlm} proposed LayoutLM method, inspired by BERT~\cite{Devlin2019BERTPO}, for document image understanding using pre-training of text and layout. Although this method uses image features and position to pre-train model and performs well on downstream tasks for document image understanding such as KIE, it doesn't consider the latent relationship between two text segments. Besides, this model needs adequate data and time consuming to pre-train model inefficiently. 

Alternative approaches~\cite{qian2018graphie,liu2019graph} predefine a graph to combine textual and visual information by using graph convolutions operation~\cite{Kipf2016SemiSupervisedCW} as illustrated in Figure~\ref{fig:compare}(c). In the literature of~\cite{qian2018graphie,liu2019graph}, the relative importance of visual features and non-sequential information is debated and graph neural networks modeling on document brings well performance on extraction entity tasks. But~\cite{qian2018graphie} needs prior knowledge and extensive human efforts to predefine task-specific \emph{edge} type and \emph{adjacent matrix} of the graph. Designing effective edge type and adjacent matrix of the graph, however, is challenging, subjective, and time-consuming, especially when the structure of documents is sophisticated.~\cite{liu2019graph} directly define a fully connected graph then uses a self-attention mechanism to define convolution on fully connected nodes. This method probably ignores the noise of the node and leads to aggregate useless and redundancy node information.

In this paper, we propose \textbf{PICK}, a robust and effective method shown in Figure~\ref{fig:compare}(d), \textbf{P}rocessing Key \textbf{I}nformation Extraction from Documents using improved Graph Learning-\textbf{C}onvolutional Networ\textbf{K}s, to improve extraction ability by automatically making full use of the textual and visual features within documents. PICK incorporates \emph{graph learning} module inspired by \cite{jiang2019semi} into existing graph architecture to learn a \emph{soft adjacent matrix} to effectively and efficiently refine the graph context structure indicating the relationship between nodes for downstream tasks instead of predefining edge type of the graph artificially. Besides, PICK make full use of features of the documents including text, image, and position features by using \emph{graph convolution} to get richer representation for KIE. The graph convolution operation has the powerful capacity of exploiting the relationship generated by the graph learning module and propagates information between nodes within a document.  The learned richer representations are finally used to a decoder to assist sequence tagging at the character level. PICK combines a graph module with the encoder-decoder framework for KIE tasks as illustrated in Figure~\ref{fig:compare}(d).

The main contributions of this paper can be summarized as follows:
\begin{itemize}
\item{In this paper, we present a novel method for KIE, which is more effective and robust in handling complex documents layout. It fully and efficiently uses features of documents (including text, position, layout, and image) to get a richer semantic representation that is crucial for extracting key information without ambiguity.} 

\item{We introduce improved graph learning module into the model, which can refine the graph structure on the complex documents instead of predefining struct of the graph.}

\item{Extensive experiments on real-world datasets have been conducted to show that our method outperforms baselines methods by significant margins.}
\end{itemize}

\section{Related Work}
Existing research recognizes the critical role played by making full use of both textual and visual features of the documents to improve the performance of KIE. The majority of methods, however, pay attention to the textual features, through various features extractors such as recurrent neural networks (RNNs) or convolutional neural networks (CNNs) on word- and character- level~\cite{lample2016neural,chiu2016named,ma2016end}. Although~\cite{Guo2019EATENEA,katti2018chargrid} uses visual image features to process extraction, it only focuses on image features and does not take textual features into account. \cite{xu2019layoutlm} attempts to use both textual and visual features for document understanding and gets good performance on some documents, through pre-training of text and layout, but it doesn't consider the relationship between text within documents. Besides, a handful of other methods~\cite{swampillai2011extracting,rusinol2013field,lebourgeois1992fast} make full use of features to support extraction tasks based on human-designed features or task-specific knowledge, which are not extensible on other documents.

Besides, recent research using both textual and visual features to aid the extraction mainly depends on graph-based representations due to graph convolutional networks (GCN)~\cite{Kipf2016SemiSupervisedCW} demonstrated huge success in unstructured data tasks. Overall, GCN methods can be split into spatial convolution and spectral convolution methods\cite{zhang2018graph}. The graph convolution our framework used to get a richer representation belongs to the spatial convolution category which generally defines graph convolution operation directly through defining an operation on node groups of neighbors~\cite{velivckovic2017graph,monti2017geometric}. Spectral methods that generally define graph convolution operation based on the spectral representation of graphs\cite{Kipf2016SemiSupervisedCW}, however, are not propitious to dynamic graph structures. \cite{song2018n,peng2017cross} proposed Graph LSTM, which enables a varied number of incoming dependencies at each memory cell. \cite{wang2018joint} jointly extract entities and relations through designing a directed graph schema. \cite{marcheggiani2017encoding} proposes a version of GCNs suited to model syntactic dependency graphs to encode sentences for semantic role labeling. \cite{gui2019lexicon} proposed a lexicon-based GCN with global semantics to avoid word ambiguities. Nevertheless, their methods don't take visual features into the model.  

The most related works to our method are~\cite{qian2018graphie,liu2019graph}, using graph module to capture non-local and multimodal features for extraction but still differ from ours in several aspects. First,~\cite{qian2018graphie} only use textual and position features where images are not used and need to predefine task-specific edge type and connectivity between nodes of the graph. Nevertheless, our method can automatically learn the relationship between nodes by graph learning module, using it to efficiently refine the structure of the graph without any prior knowledge in order to aggregate more useful information by graph convolution. Second,~\cite{liu2019graph} also does not use images features to improve the performance of extraction tasks without ambiguity. Meanwhile, due to~\cite{liu2019graph} simply and roughly regards graph as fully connectivity no matter how complicated the documents are, graph convolution aggregates useless and redundancy information between nodes. Our method, however, incorporating graph learning into framework, can filter useless nodes and be robust to document complex layout structure.

\section{Method}
\begin{figure*}[tb]
\centering
\includegraphics[width=0.99\textwidth]{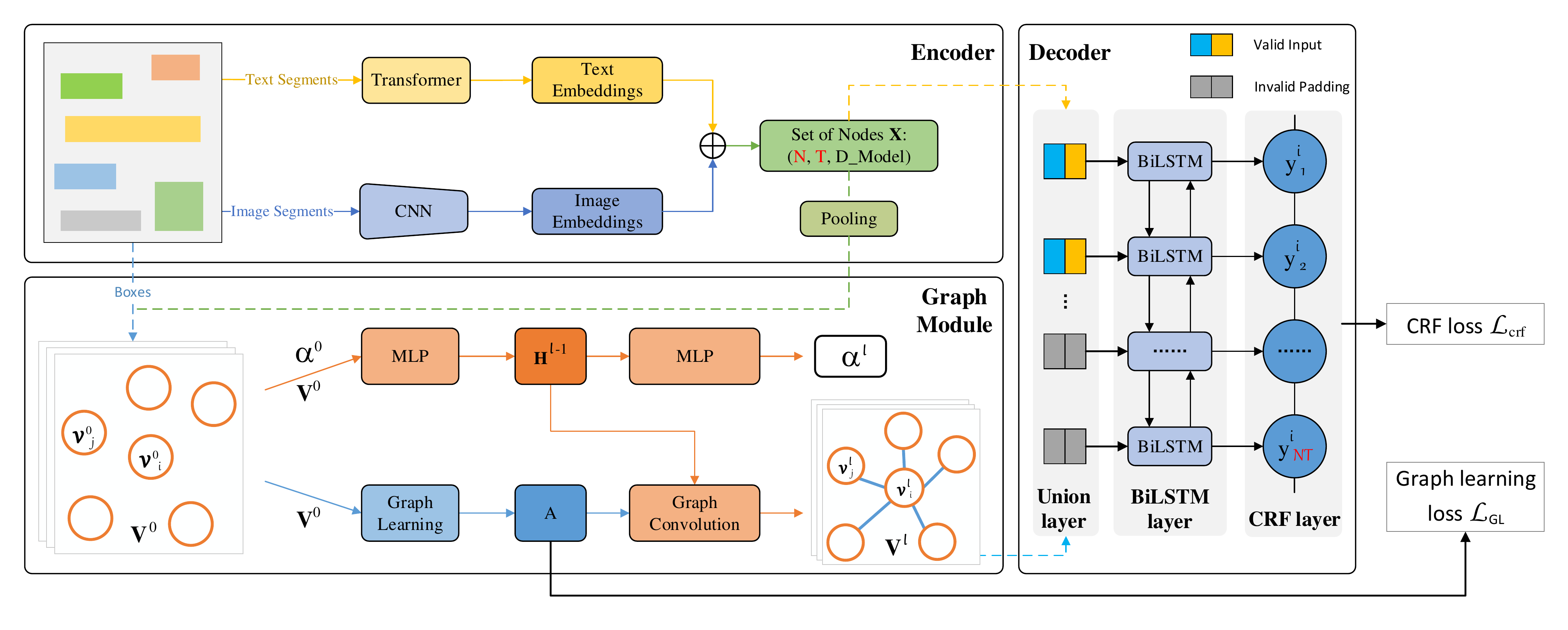}
\caption{Overview of PICK. $\textbf{V}^{l}$ donates \emph{node embedding} in the $l$-th graph convolution layer. $\bm{\alpha}^l$ and $\textbf{H}^{l}$ represents \emph{relation embedding} and \emph{hidden features} between the node $v_i$ and $v_j$ in the $l$-th graph convolution layer, respectively. $\textbf{A}$ is \emph{soft adjacent matrix}. $N$, $T$, and $D\_Model$ denotes the number of sentences segments, the max-length of sentence and the dimension of the model respectively. $\oplus$ denotes element-wise addition.}
\label{fig:overall}
\end{figure*}

In this section, we provide a detailed description of our proposed method, PICK. The overall architecture is shown in Figure~\ref{fig:overall}, which contains 3 modules: 

\begin{itemize}
\item{\emph{Encoder}}: This module encodes text segments using Transformer to get text embeddings and image segments using CNN to get image embeddings. The text segments and image segments stand for textual and morphology information individually. Then these two types of embeddings are combined into a new local representation $\mathbf{X}$, which will be used as node input to the Graph Module. 

\item{\emph{Graph Module}}: This module can catch the latent relation between nodes and get richer graph embeddings representation of nodes through improved graph learning-convolutional operation. Meanwhile, bounding boxes containing layout context of the document are also modeled into the graph embeddings so that graph module can get non-local and non-sequential features. 

\item{\emph{Decoder}}: After obtaining the graph embeddings of the document, this module performs sequence tagging on the union non-local sentence at character-level using BiLSTM and CRF, respectively. In this way, 
our model transforms key information extraction tasks into a sequence tagging problem by considering the layout information and the global information of the document.
\end{itemize}

To ease understanding, our full model is described in parts. First, we begin by introducing the notation used in this paper in Section~\ref{sec:notation}.  Our encoder representation is described in Section~\ref{sec:encoder} and
the proposed graph module mechanism is then described in Section~\ref{sec:graphmodule}. Finally, Section~\ref{sec:decoder} shows how to
combine the graph embedding and text embedding to output results.

\subsection{Notation}
\label{sec:notation}
Given a document $\mathcal{D}$ with $N$ sentences/text segments, its representation is denoted by $S=\left\lbrace s_1,\ldots, s_N \right\rbrace$, where $s_i$ is a set of characters for the $i$-th sentence/text segments. We denote $s^{is}_i$ and $s^{bb}_i$ as image segments and bounding box at position $i$, respectively. For each sentence $s_i=( c^{(i)}_1,\ldots, c^{(i)}_T)$, we label each character as $ \mathbf{y}_i=(y^{(i)}_1,\ldots, y^{(i)}_T) $ sequentially using the IOB~(Inside, Outside, Begin) tagging scheme~\cite{sang1999representing}, where $T$ is the length of sentence $s_i$. 

An accessory graph of a document $\mathcal{D}$ is denoted with $G=(V,R, E)$, where $V=\left\lbrace v_1,\ldots, v_N \right\rbrace$ is a set of $N$ nodes, $R=\left\lbrace \alpha_{i1},\ldots, \alpha_{ij} \right\rbrace$, $\alpha_{ij}$ is the set of relations between two nodes, and $E\subset  V \times R \times V$ is the edge set and each edge $e_{ij}=(v_i, \alpha_{ij}, v_j) \in E$  represents that the relation $\alpha_{ij} \in R$ exist from node $v_i$ to $v_j$.

\subsection{Encoder}
\label{sec:encoder}
As shown in Figure~\ref{fig:overall}, the top-left position in the diagram is the encoder module, which contains two branches. Different from the existing key information works~\cite{qian2018graphie,liu2019graph} that only use text segments or bounding boxes, one of our key contribution in this paper is that we also use image segments simultaneously containing morphology information to improve document representations performance which can be exploited to help key information extraction tasks.

One branch of Encoder generates text embeddings using encoder of Transformer~\cite{vaswani2017attention} for capturing local textual context. Given a sentence $s_i=( c^{(i)}_1,\ldots, c^{(i)}_T)$, text embeddings of sentence $s_i$ is defined as follows
\begin{equation}
\mathbf{te}_{1:T}^{(i)}
 = \mathtt{TransformerEncoder}\left(\mathbf{c}_{1:T}^{(i)}~;  \Theta_\text{tenc}\right)\,,
\end{equation}
where $\mathbf{c}_{1:T}^{(i)}=[\mathbf{c}_1^{(i)},\ldots,\mathbf{c}_T^{(i)}]^T \in \mathbb{R}^{ T \times d_{model} } $ denotes the input sequence, $\mathbf{c}_t^{(i)}  \in \mathbb{R}^{ d_{model}}$ represents a token embedding~(e.g., Word2Vec) of each character $c^{(i)}_t$ , $d_{model}$ is the dimension of the model, $\mathbf{te}_{1:T}^{(i)}=[\mathbf{te}_1^{(i)},\ldots,\mathbf{te}_T^{(i)}]^T \in \mathbb{R}^{ T \times d_{model} }$ denotes the output sequence, $\mathbf{te}_t^{(i)} \in \mathbb{R}^{d_{model}}$ represents the encoder output of Transformer for the $i$-th character $c_t^{(i)}$, and $\Theta_\text{tenc}$ represents the encoder parameters of Transformer. Each sentence is encoded independently and we can get a document $\mathcal{D}$ text embeddings, defining it as
\begin{equation}
\mathbf{TE}
 = [\mathbf{te}_{1:T}^{(1)};\ldots;\mathbf{te}_{1:T}^{(N)}] \in \mathbb{R}^{ N\times T \times d_{model} }\,.
\end{equation} 

Another branch of Encoder generate image embedding using CNN for catching morphology information. Given a image segment $s_i^{is}$, image embeddings is defined as follows
\begin{equation}
\mathbf{ie}^{(i)}
 = \mathtt{CNN}\left(\mathbf{s_i}^{is}~;  \Theta_\text{cnn}\right)\,,
\end{equation}
where $\mathbf{s_i}^{is} \in \mathbb{R}^{ H' \times W' \times 3 } $ denotes the vector of input image segment, $H'$ and $W'$ represent the height and width of image segment $s_i^{is}$  respectively, $\mathbf{ie}^{(i)} \in \mathbb{R}^{  H \times W \times d_{model}}$ represents the output of CNN for the $i$-th image segment $s_i^{is}$, and $\Theta_\text{cnn}$ represents the CNN parameters. We implement the CNN using ResNet~\cite{He2015DeepRL} and resize image under condition $H \times W = T $ then encode each image segments individually and we can get a document $\mathcal{D}$ image embeddings, defining it as
\begin{equation}
\mathbf{IE}
 = [\mathbf{ie}^{(1)};\ldots;\mathbf{ie}^{(N)}] \in \mathbb{R}^{ N\times T \times d_{model} }\,.
\end{equation} 

Finally, we combine text embeddings $\mathbf{TE}$ and image embeddings $\mathbf{IE}$ through element-wise addition operation for feature fusion and then generate the fusion embeddings  $\mathbf{X}$ of the document $\mathcal{D}$, which can be expressed as	
\begin{equation}
\mathbf{X}
 = \mathbf{TE} +  \mathbf{IE}\,,
\end{equation} 
where $\mathbf{X} \in \mathbb{R}^{ N\times T \times d_{model} }$ represent a set of nodes of graph ant  $\mathbf{X}$ will be used as input $\mathbf{X}_0$ of Graph Module followed by pooling operation and $\mathbf{X}_0 \in \mathbb{R}^{ N \times d_{model}}$.

\subsection{Graph Module}
\label{sec:graphmodule}
Existing key information works~\cite{qian2018graphie,liu2019graph} using graph neural networks modelling global layout context and non-sequential information need prior knowledge to pre-define task-specific \emph{edge} type and \emph{adjacent matrix} of the graph. \cite{qian2018graphie} define \emph{edge}
to horizontally or vertically connected nodes/text segments that are close to each other and specify four types of \emph{adjacent matrix}~(left-to-right, right-to-left, up-to-down, and down-to-up). But this method cannot make full use of all graph nodes and excavate latent connected nodes that are far apart in the document. Although \cite{liu2019graph} use a fully connected graph that every node/text segments is connected, this operation leads to graph aggregate useless and redundancy node information.

In this way, we incorporate improved graph \emph{learning-convolutional network} inspired by \cite{jiang2019semi} into existing graph architecture to learn a \emph{soft adjacent matrix} $\mathbf{A}$ to model the graph context for downstream tasks illustrated in the lower left corner of Figure~\ref{fig:overall}.

\subsubsection{Graph Learning}
Given an input $\mathbf{V} = [\bm{v}_1,\ldots,\bm{v}_N]^T \in \mathbb{R}^{N \times d_{model}}$ of graph nodes, where $\bm{v}_i \in \mathbb{R}^{d_{model}}$ is the $i$-th node of the graph and the initial value of $\mathbf{V}$ is equal to $\mathbf{X}_0$, Graph Module generate a \emph{soft adjacent matrix} $\mathbf{A}$ that represents the pairwise relationship weight between two nodes firstly through graph learning operation, and extract features $\mathbf{H}$ for each node $v_i$ using a multi-layer perception~(MLP) networks just like \cite{liu2019graph} on input $\mathbf{V}$ and corresponding relation embedding $\bm{ \alpha }$. Then we perform graph convolution operation on features $\mathbf{H}$, propagating information between nodes and aggregate  such information into a new feature representation $\mathbf{V}'$. Mathematically, we learn a \emph{soft adjacent matrix} $\mathbf{A}$ using a single-layer neural work as
\begin{small}
\begin{equation}
\begin{cases}
\mathbf{A}_i = \mathrm{softmax} (\mathbf{e}_{i})\,, \quad i = 1, \dots, N, \quad j = 1, \dots, N\,,\\
\mathrm{e}_{ij} = \mathrm{LeakRelu}(\mathbf{w}_i^T| \bm{v}_i -  \bm{v}_j|))\,,
\end{cases}
\label{eq:learnadj}
\end{equation}
\end{small}\\
where $\textbf{w}_i \in \mathbb{R}^{d_{model}}$ is learnable weight vector. To solve the problem of gradients vanishing at training phase, we use LeakRelu instead of Relu activation function. The function $  \mathrm{softmax}(\cdot) $ is conducted on each row of $\textbf{A}$, which can guarantee that the learned \emph{soft adjacent matrix} $\textbf{A}$ can satisfy the following property
\begin{equation}\label{eq:adjproperty}
\sum\nolimits^N_{j=1} A_{ij} =1, A_{ij}\geq 0\,.
\end{equation}

We use the modified loss function based on \cite{jiang2019semi} to optimize the learnable weight vector $\textbf{w}_i$ as follows
\begin{equation}\label{eq:glloss}
\mathcal{L}_{\textrm{GL}} = \frac{1}{N^2} \sum\nolimits^N_{i,j=1}\exp(A_{ij} + \eta \|\bm{v}_i - \bm{v}_j\|^2_2 ) +\gamma \|\textbf{A}\|^2_F\,,
\end{equation}
where $\|\cdot\|_F$ represents Frobenius-Norm. Intuitively, the first item means that nodes $\bm{v}_i$ and $\bm{v}_j$  are far apart in higher dimensions encouraging a smaller weight value $A_{ij}$, and the exponential operation can enlarge this effect. Similarly, nodes that are close to each other in higher dimensional space can have a stronger connection weight. This process can prevent graph convolution aggregating information of noise node. $\eta$ is a tradeoff parameter controlling the importance of nodes of the graph. We also average the loss due to the fact that the number of nodes is dynamic on the different documents. The second item is used to control the sparsity of \emph{soft adjacent matrix} $\textbf{A}$.  $\gamma$ is a tradeoff parameter and larger $\gamma$ brings about more sparsity \emph{soft adjacent matrix} $\textbf{A}$ of graph. We use $\mathcal{L}_{\textrm{GL}}$ as a regularized term in our final loss function as shown in Eq.(\ref{eq:totalloss}) to prevent trivial solution, i.e., $\textbf{w}_i=\textbf{0}$ as discussed in \cite{jiang2019semi}.

\subsubsection{Graph Convolution}
Graph convolutional network~(GCN) is applied to capture global visual information and layout of nodes from the graph. We perform graph convolution on the \emph{node-edge-node} triplets $(v_i, \alpha_{ij}, v_j)$ as used in \cite{liu2019graph} rather than on the node $v_i$ alone.

Firstly, given an input $\textbf{V}^{0}=\textbf{X}_0 \in \mathbb{R}^{N \times d_{model}}$ as the initial layer input of the graph, initial \emph{relation embedding} $\bm{\alpha}^0_{ij}$ between the node $v_i$ and $v_j$ is formulated as follows
\begin{equation}
\bm{\alpha}_{ij}^0 = \textbf{W}^0_{\alpha}[x_{ij}, y_{ij}, \frac{w_i}{h_i},\frac{h_j}{h_i},\frac{w_j}{h_i},\frac{T_j}{T_i}]^T\,,
\end{equation}
where $\textbf{W}^0_{\alpha} \in \mathbb{R}^{d{model} \times 6}$ is learnable weight matrix.  $x_{ij}$ and $y_{ij}$ are horizontal and vertical distance between the node $v_i$ and $v_j$  respectively. $w_i$, $h_i$, $w_j$, $h_j$  are the width and height between the node $v_i$ and $v_j$ individually. $\frac{w_i}{h_i}$ are the aspect ratio of node $v_i$, and $\frac{h_j}{h_i},\frac{w_j}{h_i}$ uses the height of the node $v_i$ for normalization and has affine invariance. Different from \cite{liu2019graph}, we also use the sentences length ratio $\frac{T_j}{T_i}$ between the node $v_i$ and $v_j$. Intuitively, the length of sentence contains latent importance information. For instance, in medical invoice, the age value entity is no more than three digits usually, which plays a critical role in improving key information extraction performance. Moreover, given the length of sentence and image, model can infer rough font size of text segments, which makes \emph{relation embedding} get more richer representation.

Then we extract \emph{hidden features} $\textbf{h}_{ij}^{l}$ between the node $v_i$ and $v_j$ from the graph using the \emph{node-edge-node} triplets $(v_i, \alpha_{ij}, v_j)$ data in the $l$-th convolution layer, which is computed by
\begin{equation}
\bm{h}_{ij}^{l} = \sigma(\textbf{W}^l_{v_{i}h} \textbf{v}_i^l + \textbf{W}^l_{v_{j}h} \textbf{v}_j^l + \bm{\alpha}_{ij}^l + \textbf{b}^l )\,,
\end{equation}
where $\textbf{W}^l_{v_{i}h} ,\textbf{W}^l_{v_{j}h} \in \mathbb{R}^{d_{model} \times d_{model}}$ are the learnable weight matrices in the $l$-th convolution layer, and $\textbf{b}^l \in \mathbb{R}^{d{model}}$ is a bias parameter. $\sigma(\cdot)=\max(0,\cdot)$ is an non-linear activation function. \emph{Hidden features} $\bm{h}_{ij}^l \in \mathbb{R}^{d_{model}}$ represent the sum of visual features and the relation embedding between the node $v_i$ and $v_j$ which is critical to aggregate more richer representation for downstream task.

Finally, \emph{node embedding} $\textbf{v}_{i}^{l+1}$ aggregate information from \emph{hidden features}  $\bm{h}_{ij}^l$ using graph convolution to update node representation. As graph learning layer can get an optimal adaptive graph \emph{soft adjacent matrix} $\textbf{A}$, graph convolution layers can obtain task-specific \emph{node embedding} by conducting the layer-wise propagation rule. For node $v_i$, we have
\begin{equation}
\textbf{v}_i^{(l+1)} = \sigma( \textbf{A}_i  \textbf{h}_{i}^{l} \textbf{W}^{l})\,,
\end{equation}
where $\textbf{W}^l \in \mathbb{R}^{d_{model} \times d_{model}}$ is layer-specific learnable weight matrix in the $l$-th convolution layer, and $\textbf{v}_i^{(l+1)} \in \mathbb{R}^{d_{model}}$ donates the node embedding for node $v_i$ in the ${l+1}$-th convolution layer. After $L$ layers, we can get a contextual information $\textbf{v}_i^{L}$ containing global layout information and visual information for every node  $v_i$. Then $\textbf{v}_i^{L}$ is propagated to the decoder for tagging task.

The \emph{relation embedding } $ \bm{\alpha}^{l+1}_{ij}$ in the ${l+1}$-th convolution layer for node $v_i$ is formulated as
\begin{equation}
\bm{\alpha}^{l+1}_{ij} = \sigma( \textbf{W}^{l}_{\alpha} \textbf{h}_{ij}^l  )\,,
\end{equation}
where $\textbf{W}^l_{\alpha} \in \mathbb{R}^{d_{model} \times d_{model}}$ is layer-specific trainable weight matrix in the $l$-th convolution layer.

\subsection{Decoder}
\label{sec:decoder}
The decoder shown in Figure~\ref{fig:overall} consists of Union layer, BiLSTM~\cite{graves2005framewise} layer and CRF~\cite{lafferty2001conditional} layer for key information extraction. \emph{Union layer} receives the input $\textbf{X} \in \mathbb{R}^{ N\times T \times d_{model} } $  having variable length T  generated from Encoder, then packs padded input sequences and fill padding value at the end of sequence yielding packed sequence $\hat{\textbf{X}} \in \mathbb{R}^{ (N \cdot T) \times d_{model} }$. Packed sequence $\hat{\textbf{X}} $ can be regarded as the union non-local document representation instead of local text segments representation when performs sequence tagging using CRF. Besides, We concatenated the \emph{node embedding} of the output of Graph Module to packed sequence $\hat{\textbf{X}}$ at each timestamps. Intuitively, node embedding containing the layout of documents and contextual features as auxiliary information can improve the performance of extraction without ambiguity. BiLSTM can use both past/left and future/right context information to form the final output. The output of BiLSTM is given by
\begin{equation}
\mathbf{Z}
 = \mathtt{BiLSTM}\left(\mathbf{\hat{X}}~; \mathbf{0},  \Theta_\text{lstm}\right) \textbf{W}_z\,,
\end{equation}
where $\mathbf{Z}=[\mathbf{z}_1,\ldots,\mathbf{z}_{N \cdot T}]^{N \cdot T} \in \mathbb{R}^{ {(N \cdot T)} \times d_{output} } $ is the output of BiLSTM and denotes the scores of emissions matrix, $d_{output}$ is the number of different entity, $\mathbf{Z}_{t,j} $ represents the score of the $j$-th entity of the $t$-th character $c_t$ in packed sequence $\mathbf{\hat{X}}$, $\mathbf{0}$ means the initial hidden state and is zero, and $\Theta_\text{lstm}$ represents the BiLSTM parameters. $\mathbf{W}_{z} \in \mathbb{R}^{ d_{model} \times d_{output} }$ is the trainable weight matrix. 

Given a packed sequence $\hat{\textbf{X}}$ of predictions $\textbf{y}$, its scores can be defined as follows
\begin{equation}
s(\hat{\textbf{X}}, \mathbf{y})=\sum_{i=0}^{{N \cdot T}} T_{y_i, y_{i+1}} + \sum_{i=1}^{{N \cdot T}} \mathbf{Z}_{i, y_i}\,,
\end{equation}
where $\mathbf{ T} \in \mathbb{R}^{({N \cdot T}+2) \times ({N \cdot T}+2)}$ is the scores of transition matrix and $\textbf{y} = (y_1, \ldots, y_{N \cdot T})$. $y_0$ and $y_{{N \cdot T}+1}$ represent the `SOS' and `EOS' entity of a sentence, which means start of sequence and end of sequence respectively. $T_{i, j}$ represents the score of a transition from the entity $i$ to entity $j$.

Then the sequence CRF layer generates a family of conditional probability via a softmax for the sequence \textbf{y} given $\hat{\textbf{X}}$ as follows
\begin{equation}
p(\mathbf{y} |\hat{\textbf{X}} ) = \frac{
	e^{s(\hat{\textbf{X}}, \mathbf{y})}
}{
	\sum_{\mathbf{\widetilde{y}} \in \mathcal{Y}(\hat{\textbf{X}})} e^{s(\hat{\textbf{X}}, \mathbf{\widetilde{y}})}
}\,,
\end{equation}
where $ \mathcal{Y}(\hat{\textbf{X}})$ is all possible entity sequences for $\hat{\textbf{X}}$.

For CRF training, we minimize the negative log-likelihood estimation of the correct entity sequence and is given by
\begin{small}
\begin{equation}\label{eq:crfloss}
\begin{cases}
\mathcal{L}_{\textrm{crf}} = -~ \log( p(\mathbf{y} |\hat{\textbf{X}})) =- s(\hat{\textbf{X}}, \mathbf{y}) + Z \,,\\
Z =  \log \left( \sum_{\mathbf{\widetilde{y}} \in \mathcal{Y}(\hat{\textbf{X}})} e^{s(\hat{\textbf{X}}, \mathbf{\widetilde{y}})} \right) = \underset{{\mathbf{\widetilde{y}} \in \mathcal{Y}(\hat{\textbf{X}}) }}{\logadd}\ s(\hat{\textbf{X}}, \mathbf{\widetilde{y}}) \,.
\end{cases}
\end{equation}
\end{small}

Our model parameters of whole networks are jointly trained by minimizing the following loss function as 
\begin{equation}\label{eq:totalloss}
\mathcal{L}_{\textrm{total}} = \mathcal{L}_{\textrm{crf}} + \lambda \mathcal{L}_{\textrm{GL}}\,,
\end{equation}
where $\mathcal{L}_{\textrm{GL}}$ and $\mathcal{L}_{\textrm{crf}}$ are defined in Eq.~\ref{eq:glloss} and Eq.~\ref{eq:crfloss} individually, and $\lambda$ is a tradeoff parameter.

Decoding of CRF layer is to search the output sequence $\textbf{y}^*$ having the highest conditional probability
\begin{equation}\label{eq:decoding}
\mathbf{y}^* = \argmax_{\mathbf{\widetilde{y}} \in \mathcal{Y}(\hat{\textbf{X}})}
{p(\mathbf{\widetilde{y}} | \hat{\textbf{X}})}\,.
\end{equation}

Training~(Eq.~\ref{eq:crfloss}) and decoding~(Eq.~\ref{eq:decoding}) phase are time-consuming procedure but we can use the dynamic programming algorithm to improve speed.

\section{Experiments}

\subsection{Datasets}
\textbf{Medical Invoice} is our collected dataset containing 2,630 images. It has six key text fields including medical insurance type, Chinese capital total amount, invoice number, social security number, name, and hospital name. This dataset mainly consists of digits, English characters, and Chinese characters. An example of an anonymized medical invoice is shown in Fig.~\ref{fig:medical-invoice}. The medical invoice is variable layout datasets with the illegible text and out of position print font. For this dataset, 2,104 and 526 images are used for training and testing individually.
\\
\textbf{Train Ticket} contains 2k real images and 300k synthetic images proposed in~\cite{Guo2019EATENEA}. Every train ticket has eight key text fields including ticket number, starting station, train number, destination station, date, ticket rates, seat category, and name. This dataset mainly consists of digits, English characters, and Chinese characters. An example of an anonymized train ticket image is shown Fig.~\ref{fig:eaten-trainticket}. The train ticket is fixed layout dataset, however, it contains background noise and imaging distortions. The datasets do not provide text bounding boxes (bbox) and the transcript of each text bbox. So we randomly selected 400 real images and 1,530 synthetic images then human annotate bbox and use the OCR system to get the transcript of each text bbox. 
For the annotated dataset, all our selected synthetic images and 320 real images are used for training and the rest of real images for testing.
\\
\textbf{SROIE}~\cite{huang2019icdar2019} contains 626 receipts for training and 347 receipts for testing. Every receipt has four key text fields consisting of company, address, date, and total. This dataset mainly contains digits and English characters. An example receipt is shown in Fig.~\ref{fig:sroie-invoice}, and this dataset have a variable layout with a complex structure. The SROIE dataset provides text bbox and the transcript of each text bbox.

\begin{table*}[htbp]
\centering
\caption{Performance comparison between PICK (Ours) and baseline method on Medical invoice datasets. PICK is more accurate than the baseline method. \textbf{Bold} represent the best performance.}
\begin{tabular}{| c | c | c | c | c | c | c |}
\hline
\multirow{2}[1]{*}{\centering \textbf{Entities} } & \multicolumn{3}{c|}{\textbf{Baseline}} & \multicolumn{3}{ c |}{\textbf{PICK (Our)}} \\
\cline{2-7}
   & \makecell*[c]{ \textbf{mEP} }  & \makecell*[c]{ \textbf{mER} }  & \makecell*[c]{ \textbf{mEF} }  &  \makecell*[c]{ \textbf{mEP} }  &\makecell*[c]{ \textbf{mER} }  & \makecell*[c]{ \textbf{mEF} }  \\
\hline

	Medical Insurance Type      &	66.8 &	77.1 &	 71.6 & 85.0    &	 81.1  &  \textbf{83.0}  \\
 
Chinese Capital Total Amount     &	85.7 &	88.9  &	87.3  &	93.1    &	98.4 &	\textbf{95.6} \\
Invoice Number    	   &	61.1 &	57.7 &	59.3 &	93.9   &	 90.9  &	 \textbf{92.4} \\
Social Security Number  &	53.4 &	64.6 &	 58.5	&	71.3 &	64.6  &	\textbf{67.8} \\
Name     &	73.1  & 73.1 &	73.1 &	74.7 &	85.6 &	 \textbf{79.8}  \\
Hospital Name      &	69.3 &	74.4 &	71.8 &	78.1 &	89.9 & \textbf{83.6} \\
\hline
Overall (micro)  & 71.1  & 73.4  &	72.3  &	85.0 &	89.2 &	\textbf{87.0} \\
\hline
   
\end{tabular}
\label{tab:medical}
\end{table*}

\subsection{Implementation Details}
\textbf{Networks Setting} In the encoder part, the text segments feature extractor is implemented by the encoder module of Transformer~\cite{vaswani2017attention} yielding text embeddings and the image segments feature extractor is implemented by ResNet50~\cite{He2015DeepRL} generating image embeddings. The hyper-parameter of Transformer used in our paper is same as~\cite{vaswani2017attention} produced outputs of dimension $d_{model} = 512$. Then the text embeddings and image embeddings are combined by element-wise addition operation for feature fusion and then as the input of the graph module and decoder. The graph module of the model consists of graph learning and graph convolution. In our experiments, the default value of $\eta, \gamma$ in graph learning loss is $1, 0.4$ individually and the number of the layer of graph convolution $L = 1$. The decoder is composed of BiLSTM~\cite{graves2005framewise} and CRF~\cite{lafferty2001conditional} layers. In the BiLSTM layer, the hidden size is set to 512, and the number of recurrent layers is 2. The tradeoff parameter of training loss $\lambda$ is 0.01 in the decoder.
\\
\textbf{Evaluation Metrics} In the medical invoice, train ticket, and SROIE scenario, in the condition of a variable number of appeared entity, mean entity recall (mER), mean entity precision (mEP), and
mean entity F-1 (mEF) defined in~\cite{Guo2019EATENEA} are used to benchmark performance of PICK.
\\
\textbf{Label Generation} For train ticket datasets, we annotated the bbox and label the pre-defined entity type for each bbox then use the OCR system to generate the transcripts corresponding to bbox. When we get bbox and the corresponding entity type and transcripts of bbox, we enumerate all transcripts of the bbox and convert it to IOB format~\cite{sang1999representing} used to minimize CRF loss. For SROIE datasets, due to the fact that it only
provides bbox and transcripts, so we annotated the entity type for each bbox, then the rest of the operation is the same as the train ticket. Different from train ticket and SROIE datasets that directly use human-annotated entity type to generate IOB format label, we adopted a heuristic approach provided in~\cite{liu2019graph} to get the label of Medical Invoice because human-annotated bbox probably cannot precisely match OCR system detected box in the process of practical application.
\\
\textbf{Implementation} The proposed model is implemented in PyTorch and trained on 8 NVIDIA Tesla V100 GPUs with 128 GB memory. Our model is trained from scratch using Adam [49] as the optimizer to minimize the CRF loss and graph learning loss jointly and the batch size is 16 at the training phase. The learning rate is set to $10^{-4}$ over the whole training phase. We also use dropout with a ratio of 0.1 on both BiLSTM and the encoder of the Transformer. Our model is trained for 30 epochs, each epoch takes about 35 minutes. At the inference phase, the model directly predicts every text segment which belongs to the most possible entity type without any post-processed operation or constraint rules to correct the results except for SROIE. For the task of extraction of SROIE, we use a lexicon which is built from the train data to autocorrect results.
\\
\textbf{Baseline method} To verify the performance of our proposed method, we apply a two-layer BiLSTM with a CRF tagger to the baseline method. This architecture has been extensively proved and demonstrated to be valid in previous work on KIE~\cite{qian2018graphie,liu2019graph}. All text segments of documents are concatenated from left to right and from top to bottom yielding a one-dimensional textual context as the input of the baseline to execute extraction tasks. The hyper-parameter of BiLSTM of the baseline is similar to the PICK method.

\begin{table}[htbp]
\caption{Results comparison on SROIE and train ticket datasets. The evaluation metric is mEF.}
\begin{center}
\small
\begin{tabular}{|c|c|c|}
\hline
  \textbf{Method }&  \textbf{Train Ticket}   &  \textbf{SROIE}  \\
\hline
Baseline & 85.4  & - \\
LayoutLM~\cite{xu2019layoutlm} & -  & 95.2  \\
\hline
PICK (Ours) & 98.6   & 96.1 \\
\hline
\end{tabular}
\end{center}
\vspace{-.5em}
\label{tab:sroieandtrain}
\end{table}
\vspace{-2em}
\begin{table}[htbp]
\caption{Results of each component of our model. The evaluation metric is mEF.}
\begin{center}
\small
\begin{tabular}{|c|c|c|c|}
\hline
  \textbf{Model }&  \textbf{Medical Invoice}   &  \textbf{Train Ticket}    \\
\hline
PICK (Full model) & 87.0  & 98.6  \\
\hline
w/o image segments &  $\downarrow$0.9 & $\downarrow$0.4 \\
w/o graph learning &  $\downarrow$1.6 & $\downarrow$0.7  \\
\hline
\end{tabular}
\end{center}
\vspace{-.5em}
\label{tab:ablationstudy}
\end{table}
\vspace{-1em}
\subsection{Experimental Results}
We report our experimental results in this section. In the medical invoice scenario, as can be seen from the Table~\ref{tab:medical}, the average mEF scores of baseline and PICK were compared to verify the performance of the PICK. What is striking about the figures in this table is that PICK outperforms the baseline in all entities, and achieves 14.7\% improvement in the overall mEF score. Further analysis showed that the most striking aspect of the data is the biggest increase in Invoice Number mEF performance. Note that Invoice Number has distinguishing visual features with red color fonts than other text segments as shown in the top left corner of Fig.~\ref{fig:medical-invoice}. In summary, these results show that the benefits of using both visual features and layout structure in KIE.

Furthermore, see from the second column of Table~\ref{tab:sroieandtrain}, PICK shows significant improvement over the baseline method in the train ticket scenario. Surprisingly, the mEF of PICK almost get a full score on the train ticket. This result suggests that PICK can handle very well extraction tasks on fixed layout documents due to PICK having the ability to learn the graph structure of documents. We also use online evaluation tools\footnote{\url{https://rrc.cvc.uab.es/?ch=13&com=evaluation&task=3}} on SROIE datasets to verify our competitive performance. As we can see from the third column of Table~\ref{tab:sroieandtrain}, our model achieves competitive results in mEF metrics in condition of only using the training data provided by official. Note that LayoutLM~\cite{xu2019layoutlm} also uses extra pre-training datasets and documents class supervised information to train the model. Data from this table can be compared with the data in Table~\ref{tab:medical} which shows the robustness of our model on both variable and fixed layout documents.

\subsection{Ablation Studies}
In order to evaluate the contributions of each component of our model, we perform ablation studies in this section. As described in Table~\ref{tab:ablationstudy}, when we remove image segments element from PICK, the most striking observation to emerge from the data comparison is the drop in performance of PICK on both medical invoice and train ticket datasets. This indicates that visual features can play an important role in addressing the issue of ambiguously extracting key information. This result is not counter-intuitive as image segments can provide richer appearance and semantic features such as font colors, background, and directions. Additionally, image segments can help the graph module capture a reasonable graph structure. Furthermore, the improved graph learning module also makes a difference in the performance of the PICK. More specifically, as shown in Table~\ref{tab:ablationstudy}, removing graph learning element from PICK leads to a large metrics score cut down on two datasets, especially on variable layout datasets. Thus, graph learning can deal with not only the fixed layout but also variable layout datasets. So graph learning element is good at dealing with the complex structures of documents and generalization.

\begin{table}[htbp]
\caption{Performance comparisons of different graph convolution layers for different datasets.The evaluation metric is mEF.}
\begin{center}
\small
\begin{tabular}{|c|c|c|c|}
\hline
  \textbf{Configuration }&  \textbf{Medical Invoice}   &  \textbf{Train Ticket}   \\
\hline
$L=1$ & 87.0  & \textbf{98.6}    \\
$L=2$ & \textbf{87.1}  & 97.2    \\
$L=3$ &  85.9  & 96.5     \\
$L=4$ &  85.34  & 92.8    \\
\hline
\end{tabular}
\end{center}
\vspace{-.5em}
\label{tab:differentlayers}
\end{table}

We perform another ablation studies to analyze the impact of the different number of layers $L$ of graph convolution on the extraction performance.  As shown in Table~\ref{tab:differentlayers}, all best results are obtained with a 1- or 2-layer model rather than a 3- or 4-layer model. This result is somewhat counter-intuitive but this phenomenon illustrates a characteristic of the GCN~\cite{Kipf2016SemiSupervisedCW} that the deeper the model (number of layers) is, the more it probably will be overfitting. In practice, we should set a task-specific number of layers of the graph.

\section{Conclusions}
In this paper, we study the problem of how to improve KIE ability by automatically making full use of the textual and visual features within documents. We introduce the improved graph learning module into the model to refine the graph structure on the complex documents given visually rich context. It shows superior performance in all the scenarios and shows the capacity of KIE from documents with variable or fixed layout. This study provides a new perspective on structural information extraction from documents.


\bibliographystyle{IEEEtran}
\bibliography{egbib}

\end{document}